\documentclass[conference]{IEEEtran}
\ifCLASSINFOpdf
\usepackage[dvipdfmx]{graphicx}
\else
\usepackage[dvipdfmx]{graphicx}
\fi
\usepackage[cmex10]{amsmath}
\usepackage{url}
\usepackage{algorithm}
\usepackage{algpseudocode}

\usepackage[dvipdfmx, usenames]{color}
\usepackage{amsfonts}
\usepackage{amssymb}
\renewcommand{\algorithmicrequire}{\textbf{Input:}}
\renewcommand{\algorithmicensure}{\textbf{Output:}}

\newcommand{\alref}[1]{Algorithm \ref{#1}}
\newcommand{\bs}[0]{\boldsymbol}
\newcommand{\paren}[1]{\left( #1 \right)}
\newcommand{\set}[1]{\left\{\, #1 \,\right\}}
\newcommand{\norm}[1]{\left\| #1 \right\|}
\newcommand{\card}[1]{\left| #1 \right|}
\newcommand{\R}[0]{\mathbb{R}}
\newcommand{\N}[0]{\mathbb{N}}
\newcommand{\Beta}[0]{\mathrm{B}}
\newcommand{\x}[0]{\bs{x}}
\newcommand{\y}[0]{\bs{y}}
\newcommand{\hf}[0]{\hat{f}}
\newcommand{\diff}[0]{\mathrm{d}}
\newcommand{\ul}[1]{\underline{#1}}
\newcommand{\T}[0]{\top}
\DeclareMathOperator*{\argmax}{argmax}
\DeclareMathOperator{\C}{C}
\DeclareMathOperator{\E}{E}

\newcommand{\mpg}[2][1.0]{
\begin{minipage}{#1 \hsize}
\begin{center}
#2
\end{center}
\end{minipage}
}

\newcommand{\ig}[2][1.0]{
\mpg[#1]{\includegraphics[width=\hsize]{#2.eps}}
}

\begin{document}
\title{Population Synthesis via $k$-Nearest Neighbor Crossover Kernel}

\author{\IEEEauthorblockN{Naoki Hamada, Katsumi Homma, Hiroyuki Higuchi and Hideyuki Kikuchi}
\IEEEauthorblockA{Fujitsu Laboratories Ltd.\\
Kanagawa, Japan\\
Email: \{hamada-naoki, km.homma, h-higuchi, h.kikuchi\}@jp.fujitsu.com}
}


%Telephone: (800) 555--1212, Fax: (888) 555--1212}


\maketitle

\begin{abstract}
%\boldmath
The recent development of multi-agent simulations brings about a need for population synthesis.
It is a task of reconstructing the entire population from a sampling survey of limited size ($1\%$ or so), supplying the initial conditions from which simulations begin.
This paper presents a new kernel density estimator for this task.
Our method is an analogue of the classical Breiman-Meisel-Purcell estimator, but employs novel techniques that harness the huge degree of freedom which is required to model high-dimensional nonlinearly correlated datasets: the crossover kernel, the $k$-nearest neighbor restriction of the kernel construction set and the bagging of kernels.
The performance as a statistical estimator is examined through real and synthetic datasets.
We provide an ``optimization-free'' parameter selection rule for our method, a theory of how our method works and a computational cost analysis.
To demonstrate the usefulness as a population synthesizer, our method is applied to a household synthesis task for an urban micro-simulator.
\end{abstract}


% \fi
\IEEEpeerreviewmaketitle

\section{Introduction} \label{sec:introduction}
In recent years, increasing computational power enables us to conduct large-scale multi-agent simulations in highly public subjects, e.g., urban planning \cite{Waddell02}, transportation \cite{Milkovits10}, energy management \cite{Vytelingum10}, disaster prevention \cite{Massaguer06} and welfare engineering \cite{Endriss04}.
To carry out multi-agent simulations in such highly public subjects, we face difficulties in collecting detailed survey data that supply realistic initial conditions to simulators.
For example, modern urban micro-simulators require disaggregate socio-demographics of the study area including each and every household's residential place, family structure, car ownership and income as well as person's age, gender, job, daily activities, etc.
The complete survey of such massive and detailed information is usually impracticable for cost and privacy reasons.

If only a sampling survey is available, we have to recover the entire population from the obtained sample.
For this purpose, Iterative Proportional Fitting (IPF) \cite{Deming40} and its extensions such as Iterative Proportional Updating (IPU) \cite{Ye09} are widely used in the aforementioned areas.
However, the IPF-like approach that simply weights, or copies, the sample points to synthesize the population cannot reproduce the diversity of the original population which might be missed through the sampling survey.
Moreover, since this approach only accepts categorical variables, numerical variables such as age and income must be roughly discretized (usually into two to five).

\begin{figure}[tb]
\ig{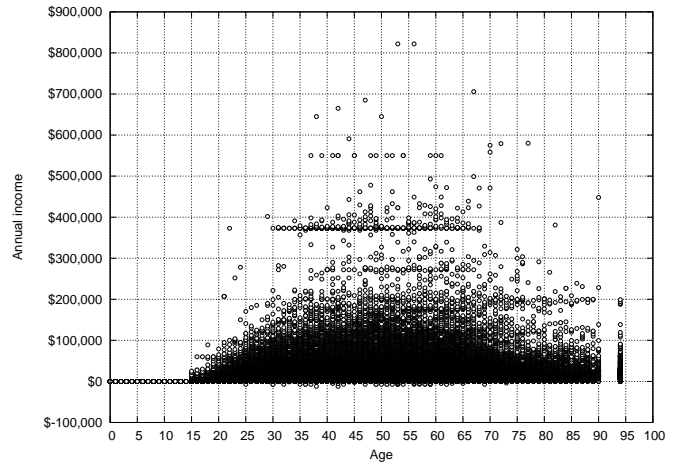}
\caption{Age vs.\ annual income of $1\%$ Washington state citizens from the American Community Survey 2012 \cite{USCB12}. Note that values at very high ages or incomes are ``top-coded'' in this survey in order to protect the confidentiality of survey respondents.}
\label{fig:age-income}
\end{figure}
One promising alternative is the statistical approach that first estimates the probability distribution of sample points and then resamples the population from the estimated distribution, though it is not easy to model complicated distributions of data for multi-agent simulations.
Figure \ref{fig:age-income} shows the scatter plot of person's age and income in a real city, which are commonly used attributes of person agents in simulations.
We can see that even this very simplified example defies parametric models.
The distribution has an income-degenerate part for under $15$ non-working ages while high income earners in middle ages constitute a one-sided long tail. Furthermore, some artifact lines arise due to ``top-coding''.
Our problem is much harder.
The data often consist of $5$ to $20$ variables and these variables are nonlinearly correlated all together.
Parametric modeling is a formidable task even for domain experts.

The kernel density estimation has been shown to possess excellent flexibility to fit various complex distributions in low-dimensional spaces \cite{Sain02}.
``Fitting'' in this method is to select the \textit{bandwidth} parameters of kernel functions so as to minimize some goodness-of-fit measure.
In high dimensions, this optimization-intensive fitting strategy becomes ineffective.
Each sample point requires a different bandwidth matrix for representing a local correlation which varies with locations in a nonlinear manner.
Due to the large number of parameters in the model, the fitting procedure involves a large-scale nonlinear optimization that requires huge computational efforts and limits the applicable dimensionality and shape of the data distribution.

Recently, a new kind of kernels for density estimation, the \textit{crossover kernel}, has been proposed \cite{Sakuma05}, which employs a crossover operator in genetic algorithm as a kernel function in kernel density estimation.
The crossover kernel does not require bandwidth parameters to specify the shape of the function; instead, it uses a subset of sample points called the \textit{kernel construction set (KCS)}.
Hence it has a potential to obviate costly fitting procedures and to quickly resample high-dimensional data.
Unfortunately, there are merely two studies on how to choose KCSs.
One appeared in Sakuma \& Kobayashi's original work: they proposed an EM algorithm to calculate sample's weights that describe the probability of which sample point belongs to which KCS.
Another is Kimura \& Matsumura's \cite{Kimura08}: they pointed out that the above method does not fit non-Gaussian-mixture-like distributions no matter how their weights are optimized, and proposed optimizing the choice of KCSs itself rather than the weights.
Both methods rely on optimization, and thus involve heavy fitting procedures again.

The aim of this paper is to develop an ``optimization-free'' crossover kernel for fast and accurate resampling by introducing the $k$-nearest neighbor restriction of KCS choice.
We will show that this simple idea works surprisingly well for our problem and has a certain theoretical background.
Our contributions are summarized as follows.
\begin{itemize}
 \item We proposed the $k$-nearest neighbor restriction of KCS choice, and showed that our method is not only faster, but also more accurate than conventional Gaussian kernels and Kimura-Matsumura crossover kernel through complicated datasets of $2$ to $17$ dimensions.
 \item We examined the parameter sensitivity of our method, and gave a rule of thumb for choosing the neighborhood size $k$ and the KCS size $m$ without optimization.
 \item We found that setting $m \le k$ decreases the generalization error in experiments, and showed its rationale from the viewpoint of  bagging \cite{Breiman96}.
 \item We demonstrated that the simulation accuracy of UrbanSim \cite{Waddell02} can be enhanced if our method supplies initial households instead of IPU.
\end{itemize}

\section{Problem Definition, Notations and Organization of the Paper} \label{sec:definition}
Throughout this paper, the following assumptions are made.
A sample $X = \set{\x_1, \dots, \x_n \in \R^d}$ of size $n (\in \N)$ is given, which is taken from an unknown population.
The population is a set of $l (\in \N)$ i.i.d.\ points with an unknown continuous density $f(\x)$.
We are going to estimate $f$, say $\hf$, and resample $l$ points from $\hf$ to recover the population.
The problem of population synthesis will be discussed under two different situations below.

\paragraph*{Case I -- an unbiased sample}
The sample $X$ is uniformly drawn from the population.
The goal is thus to estimate the density of $X$ which is identical to $f$.
The subsequent three sections of this paper discuss this case.
Section \ref{sec:conventional method} reviews existing kernel density estimators and crossover kernels.
Section \ref{sec:proposed method} presents our proposal and its properties.
Section \ref{sec:coverage} shows the effectiveness of proposed method by comparing to conventional kernel density estimators on some benchmark datasets.
Mechanisms behind the method are discussed.

\paragraph*{Case II -- a biased sample with marginal frequencies}
In real-world applications, samples available are often biased rather than uniform.
This case requires extra information about what bias is introduced.
We assume the population's binned frequencies $F = \set{f_b \in \N \mid b \in B}$ is given, but the binning $B$ is applied to marginal distributions of variables and incomplete to recover the entire joint distribution.
Thus, what we have to do in this case is to correct the sampling bias with the frequencies $F$ and to extract the variable correlation from the sample $X$, combining them to estimate the entire joint density $f$.
In Section \ref{sec:application}, we will tackle an application of this type and will present a bias correction technique.
Incorporated with our resampling method developed in Section \ref{sec:proposed method}, this technique is applied to a task of generating initial population of households for an urban land-use micro-simulation.

\section{Conventional Methods} \label{sec:conventional method}
Now focusing on Case I, this section reviews existing kernel density estimators and crossover kernels.
\subsection{Kernel Density Estimation} \label{sec:KDE}
Kernel density estimators can be classified into two versions \cite{Jones90}: the \textit{sample point estimator} and the \textit{balloon estimator}.
For resampling purpose, our estimate must be a density and easy to sample.
The sample point estimator satisfies both requirements if the kernel is a density which is easy to sample, while the balloon estimator is usually not a density even when the kernel is \cite{Terrell92}.
Thus we concentrate on the former of the following form:
\begin{equation} \label{eqn:KDE}
\hf(\y) = \frac{1}{n} \sum_{i=1}^n K(\y | \x_i, \bs H_i)
\end{equation}
where $K$ is a \textit{kernel function} to be specified and $\bs H_i$ is a \textit{bandwidth matrix} that is a $d \times d$ symmetric positive-definite matrix of parameters that we are estimating.
In this paper, we consider the multivariate Gaussian kernel:
\begin{multline} \label{eqn:Gaussian kernel}
K(\y | \x_i, \bs H_i) = \mathrm{N}(\y | \x_i, \bs H_i) = \frac{1}{\sqrt{(2 \pi)^{d} \det(\bs H_i)}}\\
\cdot \exp \paren{-\frac{1}{2}(\y - \x_i)^\T \bs H_i^{-1} (\y - \x_i)}.
\end{multline}

As a measure of goodness-of-fit, the \textit{mean integrated squared error (MISE)} is most commonly used\footnote{Various measures based on the $L_1$-errors, the Kullback-Leibler divergence and the Hellinger distance are used for this purpose \cite{Simonoff96}. In any case, the problem that we will point out in this section still arises.} and its bias/variance decomposition is given by
\begin{equation} \label{eqn:MISE}
\begin{split}
\mathrm{MISE} & = \E \left [ \int \paren{\hf(\y) - f(\y)}^2 \diff \y \right ]\\
                     & = \int \underbrace{\paren{\E[\hf(\y)] - f(\y)}^2}_{\mbox{squared bias}} \diff \y\\
                     &\quad + \int \underbrace{\E \left[ \paren{\hf(\y) - \E[\hf(\y)]}^2 \right]}_{\mbox{variance}} \diff \y.
\end{split}
\end{equation}
In the case of fixed matrix bandwidths ($\bs H_1 = \dots = \bs H_n$), bandwidth selectors using cross-validation \cite{Duong05} and plug-in \cite{Chacon10} have been developed.
However, when it comes to variable bandwidths, existing techniques are very restrictive.
The Breiman-Meisel-Purcell (BMP) estimator \cite{Breiman77} is the most classical one, which uses scalar bandwidths, i.e., $\bs H_i = h_i^2 \bs I$, and determines $h_i$ as a constant multiple of the Euclidean distance $\delta_{ik}$ from $\x_i$ to the $k$th nearest other sample point:
\begin{equation} \label{eqn:BMP}
h_i = h \delta_{ik} \quad \mbox{for fixed } h > 0.
\end{equation}
Asymptotically, this is equivalent to choosing $h_i \propto f(\x_i)^{-1/d}$ as $n \to \infty$ \cite{Terrell92}.
Abramson \cite{Abramson82} proposed the square root law, i.e., choosing $h_i \propto f(\x_i)^{-1/2}$ regardless of $d$, and showed that it reduces bias and accelerates the rate of convergence.
To handle matrix bandwidths, Sain \cite{Sain02} studied a binning technique which decreases the number of parameters by sharing the same bandwidth matrix with sample points in the same bin.
His simulation results clarified that even for a low-dimensional ($d=2$) moderate sample size ($n=320$), the unbiased cross-validation tends to yield too small, namely, overfit bandwidth matrices.
Since the (unbinned) variable bandwidth matrix has $nd(d+1)/2$ parameters, the situation is much worse in higher dimensions.
Developing bandwidth selectors for this case is currently an open question.

Another problem in high dimensions is that the relative contribution of variance dominates bias in MISE.
Sain \cite{Sain02} pointed out that for the fixed scalar bandwidth, the ratio of order of bias to order of variance is $4:d$.
Therefore, as dimensionality increases, bias reduction techniques get less importance and variance reduction techniques including the bagging \cite{Breiman96} become the core of the estimator.

\subsection{Crossover Kernels} \label{sec:crossover kernel}
In the genetic algorithm literature, Kita et al.\ \cite{Kita99} argued the desirable behavior of crossover operators and suggested a guideline, the \textit{preservation of statistics}, which tells us that the crossover's parents and children should have the same mean and covariance.
Specifically, suppose that the parents and children are realizations of random variate $\x$ and $\y$ in $\R^d$, respectively, and then the following should be satisfied:
\begin{equation} \label{eqn:PoS}
\E[\x] = \E[\y],\qquad
\C[\x] = \C[\y].
\end{equation}

Sakuma \& Kobayashi \cite{Sakuma05} pointed out that the parents $\x_1, \dots, \x_m$ (realizations of $\x$) implicitly determine the probability density function $K$ of $\y$ and thus $K$ can be considered as a data-driven kernel function.
In this context, the set of parents $X' = \set{\x_1, \dots, \x_m}$ is called the \textit{kernel construction set (KCS)}.
If one chooses a crossover so that it satisfies \eqref{eqn:PoS} and $K$ is the Gaussian distribution, then the crossover coincides with the maximum likelihood estimator (MLE) of $K$ calculated from the KCS.
Sakuma \& Kobayashi also noticed that UNDX-$m$ \cite{Kita99} fulfills these conditions when the KCS size is set to be $m=d + 1$.

The remaining question is how to construct the entire density estimate from crossover kernels.
Sakuma \& Kobayashi \cite{Sakuma05} just used the Gaussian mixture model $\hf(\y) = \sum_{i=1}^k \alpha_i \mathrm{N}(\y | \bs \mu_i, \bs \Sigma_i)$ as the final estimate, but its parameters $\alpha_i, \bs \mu_i, \bs \Sigma_i$ are optimized by their modified EM algorithm in which each Gaussian component at the $t$th EM step $\mathrm{N}(\y | \bs \mu_i^{(t)}, \bs \Sigma_i^{(t)})$ is replaced by the average of crossover kernels
\begin{equation} \label{eqn:Sakuma2}
\mathrm{\widetilde{N}}(\y | X_{i1}^{(t)}, \dots, X_{iL}^{(t)}) = \frac{1}{L} \sum_{l=1}^L K(\y | X_{il}^{(t)}).
\end{equation}
Their KCSs $X_{il}^{(t)}$ are chosen at random from $X$ by the probability proportional to weights $w^{(t)}(i | \x_j)$ which are calculated by their E step and describe the possibility that the $i$th component generates the $j$th sample point.
The superiority of their EM algorithm over the original one has been shown through simulation studies.

Kimura \& Matsumura \cite{Kimura08} proposed a more direct way, giving the entire estimate by
\begin{equation} \label{eqn:Kimura}
\hf(\y) = \frac{1}{L} \sum_{l=1}^L K(\y | X_l).
\end{equation}
Their KCSs $X_l$ are chosen so as to maximize the log-likelihood $\sum_{i=1}^n \log \hf(\x_i)$: starting from a random choice, KCSs are randomly remade (one at a time) if it improves the log-likelihood.
They showed the high flexibility of this approach by applying it to a circularly distributed dataset.

\section{$k$-Nearest Neighbor Crossover Kernel} \label{sec:proposed method}
As we have seen, the BMP estimator is fast but restricted to the scalar bandwidth, whereas the crossover kernel is slow but able to handle the covariance matrix.
Our idea is to combine the strengths of both estimators.

\subsection{REX Kernel}
As mentioned in Section \ref{sec:crossover kernel}, UNDX-$m$ coincides with the Gaussian MLE only when the KCS size is $m = d + 1$.
In order to extend this property to arbitrary size, we employ REX \cite{Kobayashi09} for the kernel function.
REX generates a resampling point $\y$ from the KCS $X' = \set{\x_1, \dots, \x_m}$ as follows:
\begin{equation}\label{eqn:REX}
 \y = \bs \mu_{X'} + \sum_{i = 1}^{m} \varepsilon_i (\x_i - \bs \mu_{X'}), \quad \varepsilon_i \sim \mathrm{N}(0,\, \frac{1}{m})
\end{equation}
where $\bs \mu_{X'} = \frac{1}{m} \sum_{i=1}^m \x_i$ and $\varepsilon_i$ is a random number drawn from the univariate Gaussian distribution $\mathrm{N}(0,\, \frac{1}{m})$.

Following Sakuma \& Kobayashi's \cite{Sakuma05} derivation of the kernel function of UNDX-$m$, we get the probability density function of $\y$, namely, the REX kernel:
\begin{multline}\label{eqn:REX pdf}
K(\y | X') = \mathrm{N}(\y | \bs \mu_{X'}, \bs \Sigma_{X'})
              = \frac{1}{\sqrt{(2 \pi)^{d} \det(\bs \Sigma_{X'})}}\\
\cdot \exp \paren{-\frac{1}{2}(\y - \bs \mu_{X'})^\T \bs \Sigma_{X'}^{-1} (\y - \bs \mu_{X'})}
\end{multline}
where $\bs \Sigma_{X'} = \frac{1}{m} \sum_{i=1}^m (\x_i - \bs \mu_{X'}) \otimes (\x_i - \bs \mu_{X'})^\T$.
Regardless of $m$, thus always $K(\y|X')$ is the Gaussian MLE calculated from $X'$.
Comparing the REX kernel \eqref{eqn:REX pdf} to the Gaussian kernel \eqref{eqn:Gaussian kernel}, we see that the covariance matrix $\bs \Sigma_{X'}$ in \eqref{eqn:REX pdf} is corresponding to the bandwidth matrix $\bs H_i$ in \eqref{eqn:Gaussian kernel}.
Resampling from the Gaussian kernel with explicit MLE costs $O(md^2 + d^3)$ time;\footnote{Since the random number generation relies on the Cholesky decomposition of $\bs \Sigma_{X'}$, it requires $O(md^2)$ time to calculate $\bs \Sigma_{X'}$ from $X'$ and $O(d^3)$ time the Cholesky decomposition of $\bs \Sigma_{X'}$.} the REX kernel with resampler \eqref{eqn:REX} bypasses the estimation process and enables us to generate a sample point from $K(\y | X')$ in $O(md)$ time.

\subsection{$k$-Nearest Neighbor Restriction of KCS Choice}
In order to adapt KCSs to data without optimization, we choose KCSs from the $k$-nearest neighbor ($k$-NN) of each sample point.
This is a simple yet powerful trick to realize a crossover kernel analogue of the BMP estimator.
It is shown in Appendix \ref{sec:derivation} that the covariance matrix $\bs \Sigma_i$ of $k$-NN points of a sample point $\x_i$ is asymptotically of the form:
\begin{equation} \label{eqn:k-NN MLE}
\bs \Sigma_i \approx \frac{\delta_{ik}^2}{d + 2} \bs I - \frac{\delta_{ik}^4}{(d + 2)^2} \frac{\nabla f \otimes \nabla f^\T}{f^2} (\x_i) \quad \mbox{as } n \to \infty.
\end{equation}
For sufficiently large $n$, we can expect $\delta_{ik} \ll 1$ and thus the first $O(\delta_{ik}^2)$ term in the r.h.s.\ overwhelms the second $O(\delta_{ik}^4)$ term. This means that \eqref{eqn:k-NN MLE} approximates to a scalar bandwidth of $\delta_{ik}/ \sqrt{d+2}$, that is, the BMP bandwidth \eqref{eqn:BMP} with $h = 1 / \sqrt{d+2}$.
Such an asymptotic behavior has a strength and a weakness: one hand, our method inherits rich asymptotics from the BMP estimator, including the normality, the consistency, the leading term of bias/variance, the rate of convergence, etc. \cite{Silverman86}; on the other hand, our matrix extension of bandwidth gives no advantage over the BMP estimator in asymptopia.

\begin{figure}[tb]
\ig{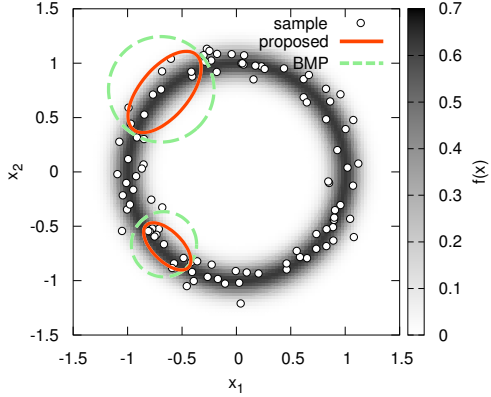}
\caption{An example of local correlation. Two typical $\bs \Sigma_i$'s of the proposed method \eqref{eqn:k-NN MLE} are shown by solid ellipsoids. The corresponding $h_i$'s of the BMP estimator \eqref{eqn:BMP} with $h = 1$ are drawn by dashed circles. The sample points (the dots) and the underlying density (the gray scale) are of the ring-like PDF \cite{Kimura08}.}
\label{fig:local correlation}
\end{figure}
For small-to-moderate $n$, the second term in the r.h.s.\ of \eqref{eqn:k-NN MLE} can be influential when nonlinear correlation presents.
Notice that the matrix-valued function $(\nabla f \otimes \nabla f^\T) / f^2$ in the second term captures the local correlation around $\x_i$, which is not taken into account in the BMP estimator.
It is easy to check that the matrix $(\nabla f \otimes \nabla f^\T)(\x)$ is always of rank one unless the gradient $\nabla f(\x)$ vanishes, and its unique nonzero eigenvalue equals $\norm{\nabla f(\x)}^2$ and the corresponding eigenvector is parallel to $\nabla f(\x)$.
Dividing $(\nabla f \otimes \nabla f^\T)(\x)$ by $f^2(\x)$ can be thought of as a relativization of the effect.
Thus, subtracting such a matrix from the BMP bandwidth as in \eqref{eqn:k-NN MLE} makes the bandwidth shrink in the gradient direction.
Figure \ref{fig:local correlation} illustrates how this effect captures the local correlation.
In this example, the gradient is everywhere normal to the circumference of the circular density and becomes very large near the circumference.
Thus, the bandwidth shrinkage does occur in the normal directions and does not in the tangent directions with respect to the density manifold, resulting in a better fit shown in Fig.\ \ref{fig:local correlation}.
Accordingly, our matrix extension of bandwidth will gain some improvement in finite sample size.

Another advantage of our method over the BMP estimator in finite sample size is that the BMP bandwidth depends on only the $k$th nearest sample point and the other $k-1$ points are ignored while our bandwidth is determined by all the $k$ points.
This will make the bandwidth computation more stable.

\subsection{Bagging of Kernels} \label{sec:bagging}
Even our $k$-NN MLE is stabler than BMP's $k$-NN distance, the severe variability of high-dimensional samples will still fluctuate the resulting kernel.
In order to exploit the sample information for further variance reduction, we employ the bagging \cite{Breiman96} of kernels.
That is, our KCS is not the $k$-NN itself, but a subset of it, and we change the KCS every time a resampling point is generated.
The resulting entire estimate becomes as follows:
\begin{equation}\label{eqn:bagging REX kernel}
 \hf(\y) = \frac{1}{n} \sum_{i=1}^n \E_{X'_i} [K(\y | X'_i)]
\end{equation}
where $\E_{X'_i}[\;\cdot\;]$ denotes the expectation over all possible choices of KCS $X'_i$ of size $m$ from the $k$-NN of $\x_i$.

Of course, we cannot compute the value of $\hf(\y)$ since $\E_{X'_i} [K(\y | X'_i)]$ is a mixture of too many Gaussians, but we can easily draw a sample from it by simply choosing $\x_i$ and its $k$-NN points at random and passing them to \eqref{eqn:REX}.
We end this section by presenting the resampling algorithm via the $k$-NN REX kernel in \alref{alg:XKDE}.
\begin{algorithm}[t!]
\footnotesize
\caption{$k$-NN REX Kernel}
\label{alg:XKDE}
\begin{algorithmic}[1]
\Require
\Statex $X = \set{\x_1, \dots, \x_n} \subset \R^d$ \Comment{Sample (training data)}
\Statex $k \in \set{0, \dots, n}$ \Comment{Neighborhood size}
\Statex $m \in \set{1, \dots, k+1}$ \Comment{KCS size}
\Statex $l \in \N$ \Comment{Population size}
\Ensure
\Statex $Y = \set{\y_1, \dots, \y_l} \subset \R^d$ \Comment{Synthetic population}
\ForAll{$\x \in X$}
\State calculate the $k$-NN of $\x$, say $X(\x, k)$
\EndFor
\State $Y \gets \emptyset$
\Repeat
\State draw $\x_1$ from $X$ at random
\State draw $\x_2, \dots, \x_m$ $(\x_i \ne \x_j)$ from $X(\x_1, k)$ at random \label{alg:XKDE KCS}
\State generate $\y$ by \eqref{eqn:REX}
\State $Y \gets Y \cup \set{\y}$
\Until{$\card{Y} = l$}
\State \Return $Y$
\end{algorithmic}
\end{algorithm}

\section{Experiments} \label{sec:coverage}
In order to evaluate the proposed method, we conduct numerical experiments on benchmark datasets. The population reproducibility, the parameter sensitivity, the bagging effect and the CPU time are studied.

\subsection{Methods and Parameter Settings} \label{sec:setting}
\textbf{$k$-NN REX kernel:} \alref{alg:XKDE} was used as the proposed method.
Its parameters were set to be as follows: the neighborhood size $k = 0, \dots, 99$, and the KCS size $m = 1, \dots, k+1$ for each $k$.

\textbf{KM REX kernel:} As a state-of-the-art crossover kernel, we combined the Kimura-Matsumura (KM) model \eqref{eqn:Kimura} with the REX kernel \eqref{eqn:REX pdf}.
Its parameters were set to be as follows: the number of KCSs $L = 1, \dots, 100$ and the KCS size $m = d + 1, \dots, d + 10$.
The optimization of KCS choice ran until consecutive $10000$ iterations do not increase the log-likelihood.

\textbf{BMP Gaussian kernel:} \eqref{eqn:KDE}, \eqref{eqn:Gaussian kernel} with $\bs H_i = h_i^2 \bs I$ and \eqref{eqn:BMP} was used to see our performance progress owing to the bandwidth matrization and the bagging.
Its parameters were set to be as follows: the neighborhood size $k = 1, \dots, 99$ and the constant multiplier $h = 0.01, \dots, 1.00$.

\textbf{Fixed Gaussian kernel:} \eqref{eqn:KDE} and \eqref{eqn:Gaussian kernel} with $\bs H_i = h^2 \bs I$ was used as a conventional estimator.\footnote{This is virtually (but, not exactly) identical to using a fixed matrix bandwidth, i.e., $\bs H_1 = \dots = \bs H_n$ in \eqref{eqn:Gaussian kernel}, because our datasets are normalized by an affine transformation.}
Its parameter, the bandwidth, was set to be $h = 0.001, \dots, 0.100$.

\subsection{Datasets} \label{sec:benchmark}
To examine the effect of dimensionality and sample size, four datasets shown in Table \ref{tbl:datasets} were used.
Every dataset was affinely transformed by the whitening \cite{Fukunaga72} so that its mean and covariance can be the zero vector and identity matrix, respectively.
All the datasets and their detailed descriptions are available online; consult the references in the URL column of the table.
\begin{table}[t]
\begin{center}
\caption{Description of datasets.}
\label{tbl:datasets}
\begin{tabular}{cccc} \hline
Name				& Dimension	& Sample Size	& URL\\
\hline
PUMS Person in WA	& $2$		& $69301$		& \cite{USCB12}\\
Swiss Roll			& $3$		& $10000$		& \cite{Gerber11}\\
$3$D Road Network	& $4$		& $434874$		& \cite{Bache13}\\
Letter Recognition		& $17$		& $20000$		& \cite{Bache13}\\
\hline
\end{tabular}
\end{center}
\end{table}

\subsection{Criterion} \label{sec:metrics}
In order to evaluate the accuracy of population reproduction, we measured the Hellinger distance \cite{Bhattacharyya43} between test data and a synthesized population.
Each dataset was evenly divided into $100$ subsets,\footnote{For datasets that cannot be divided by $100$, the remainders were randomly dropped.} and then used for the inverted cross-validation (ICV): one subset was used as training data and the remaining $99$ subsets constituted test data.
Since we cannot compute the value of the $k$-NN REX kernel's density \eqref{eqn:bagging REX kernel} as stated in Section \ref{sec:bagging}, a binned version of the Hellinger distance should be used.
For a reconstructed population $Y$ and a set of test data $Z$, our performance measure can be written by
\begin{equation}
{\rm H}(Y, Z) = \sqrt{\frac{1}{2}\sum_{b \in B} \paren{\sqrt{\card{Y_b} / \card{Y}} - \sqrt{\card{Z_b} / \card{Z}}}^2}
\end{equation}
where $B$ is the set of bins, $Y_b$ (resp. $Z_b$) is a subset of $Y$ (resp. $Z$) whose elements fall in the bin $b$, and $\card{\; \cdot \;}$ is the set cardinality.
When ${\rm H}(Y, Z) < {\rm H}(Y', Z)$ holds, the method that generates $Y$ has a better accuracy than the method that generates $Y'$.

For each setting, the $100$-fold ICV was done once, and then the average and standard deviation over its $100$ Hellinger distances were calculated.

\begin{table*}[t]
\begin{center}
\caption{Hellinger distance (average $\pm$ standard deviation over $100$-fold ICV) with best parameter settings. In each row, the bolded average is smaller than any other one with statistical significance $p < 0.01$ by Welch's $t$-test.}
\label{tbl:hellinger}
\begin{tabular}{c|ccccc} \hline
Dataset							& $k$-NN				& KM					& BMP					& Fixed					& Training Data			\\
($d$ dimensions, $n$ training points)	& REX Kernel				& REX Kernel				& Gaussian Kernel			& Gaussian Kernel			& (Baseline)				\\
\hline
PUMS Person in WA				& {\bf 2.14e-1} $\pm$ 6.82e-3	& 2.87e-1 $\pm$ 2.06e-2	& 2.46e-1 $\pm$ 9.74e-3	& 3.39e-1 $\pm$ 1.09e-2	& 5.11e-1 $\pm$ 5.73e-3	\\
$(d=2, n=693)$					& $(k=66, m=3)$			& $(L=35, m=4)$			& $(k=2, h=0.99)$			& $(h=0.011)$				&		 				\\
\hline
Swiss Roll						& {\bf 3.41e-1} $\pm$ 1.48e-2	& 3.89e-1 $\pm$ 2.63e-2	& 3.74e-1 $\pm$ 1.32e-2	& 4.02e-1 $\pm$ 1.24e-2	& 6.65e-1 $\pm$ 1.34e-2	\\
$(d=3, n=100)$					& $(k=12, m=3)$			& $(L=66, m=6)$			& $(k=22, h=0.12)$			& $(h=0.056)$				& 						\\
\hline
$3$D Road Network				& {\bf 1.95e-1} $\pm$ 1.01e-3	& 3.74e-1 $\pm$ 4.20e-3	& 1.98e-1 $\pm$ 3.33e-3	& 2.36e-1 $\pm$ 3.11e-3	& 2.78e-1 $\pm$ 3.80e-3	\\
$(d=4, n=4348)$					& $(k=30, m=2)$			& $(L=96, m=5)$			& $(k=3, h=0.18)$			& $(h=0.009)$				&		 				\\
\hline
Letter Recognition					& {\bf 5.07e-1} $\pm$ 1.00e-2	& 5.64e-1 $\pm$ 1.42e-2	& 5.45e-1 $\pm$ 1.12e-2	& 5.47e-1 $\pm$ 1.08e-2	& 7.00e-1 $\pm$ 1.26e-2	\\
$(d=17, n=200)$					& $(k=36, m=4)$			& $(L=68, m=20)$			& $(k=2, h=0.12)$			& $(h=0.059)$				& 						\\
\hline
\end{tabular}
\end{center}
\end{table*}

\subsection{Results} \label{sec:result}
Table \ref{tbl:hellinger} shows the Hellinger distances from the test data of $99n$ points to the population of $99n$ points synthesized by each method with $n$ training data.
The Hellinger distances to the training data themselves are also displayed as a baseline.
Note that their values are attainable by simply copying or bootstrapping the training data.

Every method much better reproduced the population than the baseline for all datasets, except the KM REX kernel for $3$D Road Network.
In this sense, the kernel density estimation is, more or less, generally preferable to copying and bootstrapping approaches for the purpose of population synthesis, especially when the sample is small and complex.
Among others, the $k$-NN REX kernel gained further significant improvements in all cases.
For visual comparison, we provide the scatter plots of synthetic populations in Appendix \ref{sec:visual comparison}.

A surprising fact can be found in the comparison of their standard deviations.
The $k$-NN REX kernel's standard deviations were less than the KM REX kernel's and as low as the other three cases.
Remember that our model is a matrix bandwidth extension of the BMP model and uses a lot more kernels than the KM model; thus it seems natural to expect our standard deviations have the greatest values.
That paradoxical result implies the success of our variance reduction techniques.
It holds good even in Letter Recognition, i.e., a high-dimensional ($d=17$), small sample size $(n=200)$ case.

\subsection{Discussions} \label{sec:discussion}
\subsubsection{Parameter Sensitivity} \label{sec:parameter sensitivity}
\begin{figure*}[htb]
\hspace{-8mm}
\begin{tabular}{cc}
\ig[0.5]{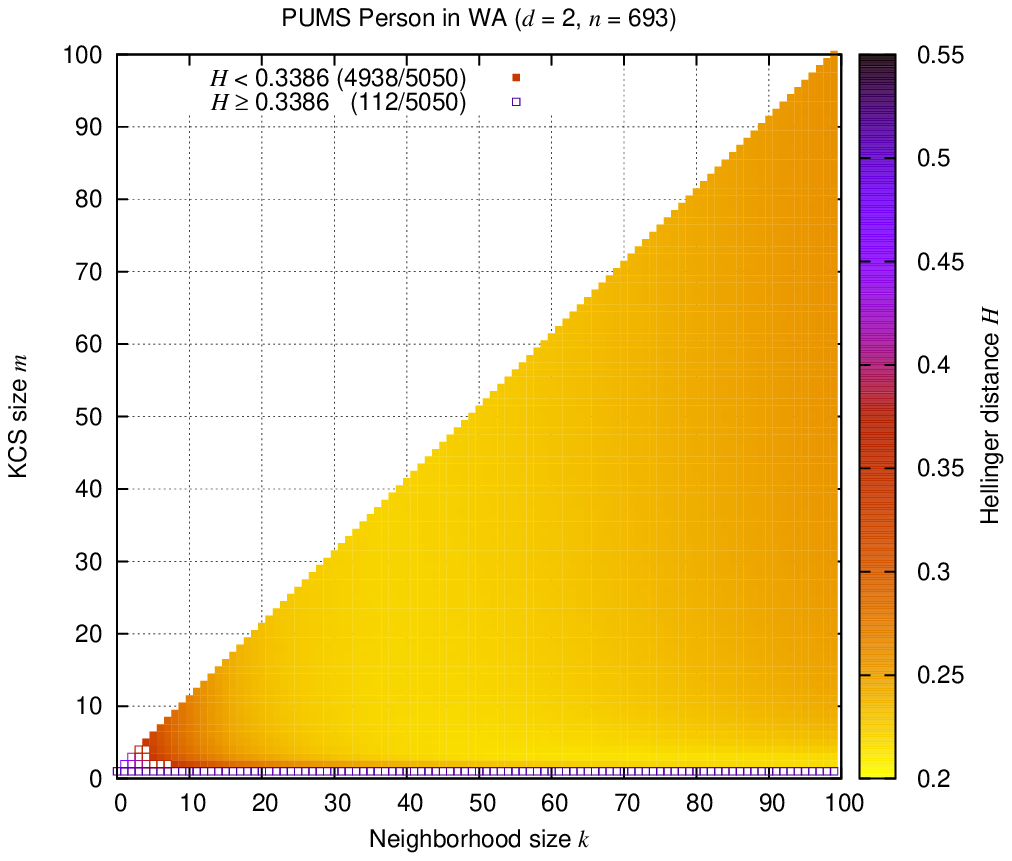}
\ig[0.5]{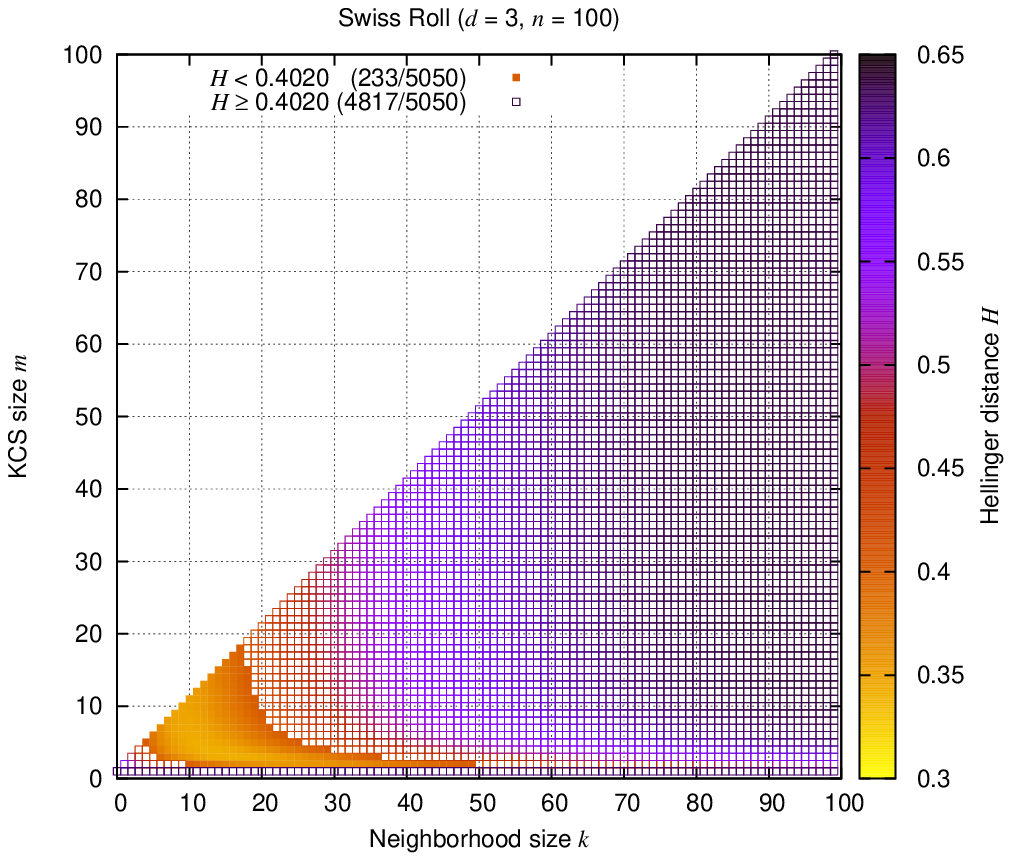}\\
\ig[0.5]{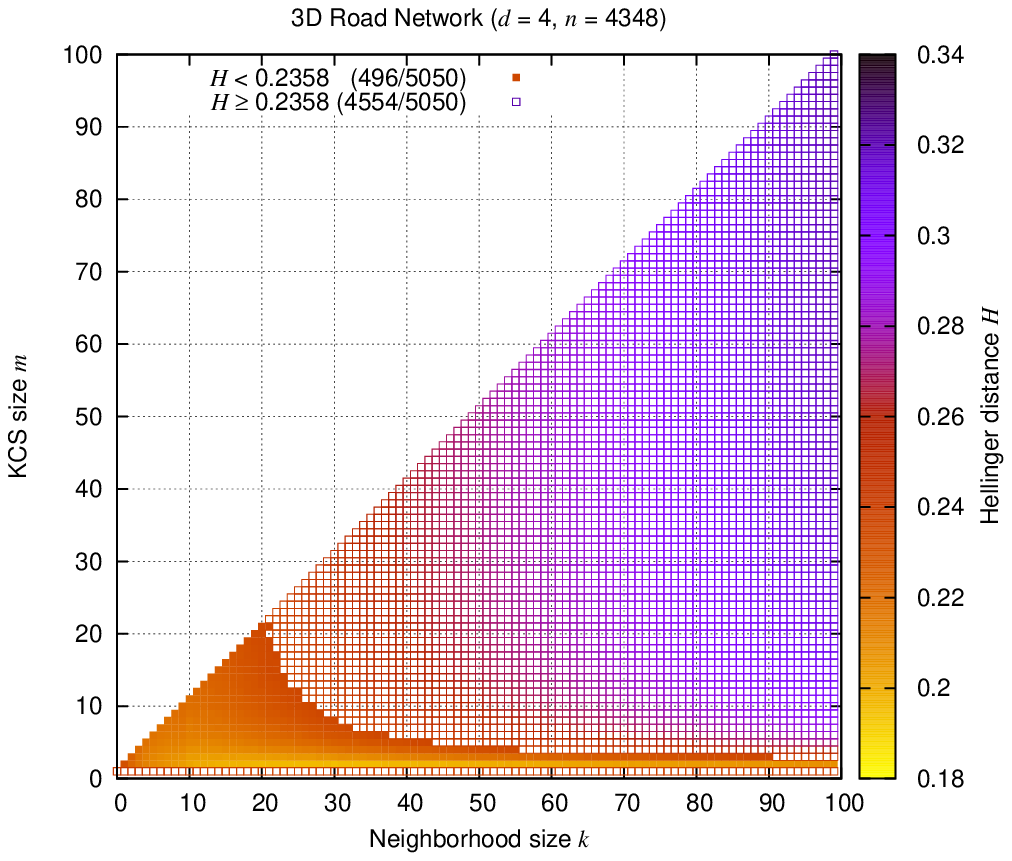}
\ig[0.5]{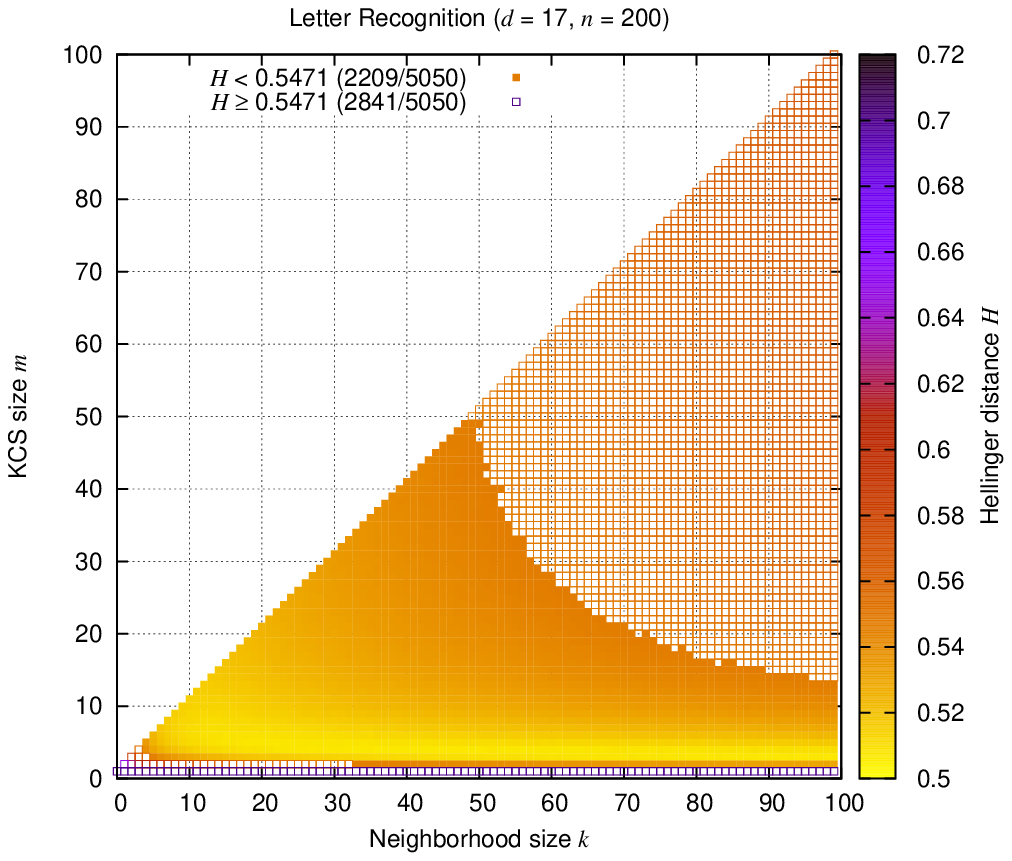}\\
\mpg{
\caption{Parameter $(k, m)$ vs.\ Hellinger distance of the $k$-NN REX kernel. The Hellinger distance is averaged over $100$-fold ICV. The legend ``$H < \alpha \ (\beta/5050) \ \blacksquare$'' means as follows: Hellinger distances at filled squares are better than the best tuned Gaussian kernel's Hellinger distance $\alpha$, and there are $\beta$ out of 5050 parameter settings that achieve better distances. The legend ``$H \ge \alpha \ (\beta/5050) \ \square$'' should be read similarly.}
\label{fig:param pc}
}
\end{tabular}
\end{figure*}

Figure \ref{fig:param pc} gives a survey of all results of the $k$-NN REX kernel.
The filled area in each plot indicates the parameters whose results surpass the best performance of the fixed Gaussian kernel.
Every filled area has a horizontally long part, which means the performance of the $k$-NN REX kernel is insensitive to the neighborhood size $k$.
Our observation is consistent with the sensitivity analysis for $k$ of the BMP estimator studied by Breiman et al.\ \cite{Breiman77}.

The figure also clarifies that the choice of the KCS size $m$ is critical to the performance.
How should practitioners find a suitable $m$?
For PUMS Person in WA ($d=2$), the optimal value is $m=3$.
This agrees with a common practice in genetic algorithm: if the search space is of $d$ dimensions, then the crossover should use $m = d + 1$ parents \cite{Kobayashi09}.
For Swiss Roll ($d=3$), $3$D Road Network ($d=4$) and Letter Recognition ($d=17$), the optimal $m$'s are $3$, $2$ and $4$, respectively.
It looks inconsistent, but once their intrinsic dimensions are considered, the same rule can be found.
Swiss Roll is a $2$-dimensional spiraling band distribution.
In $3$D Road Network, the data are observed along roads, and the intrinsic dimension is therefore $1$.
The dimensionality of Letter Recognition can be reduced to about $3$ by nonlinear dimensionality reduction methods \cite{Zhang05}.
These remind us of a practice in multiobjective genetic algorithm: when the Pareto solution set forms a $d'$-dimensional submanifold in the $d$-dimensional search space, using $m = d' + 1$ parents exhibits a better performance than $m = d  + 1$ \cite{Hamada08}.

As a consequence, we suggest a parameter selection rule that determines $k$ and $m$ by the following steps:
\begin{enumerate}
 \item Choose $k$ between $10$ and $50$.
 \item Identify the intrinsic dimensionality $d'$ of the dataset by a priori knowledge or dimensionality reduction.
 \item Set $m = d'+1$, or cross-validate near $d'$ if possible.
\end{enumerate}

\subsubsection{Effect of Bagging} \label{sec:effect bagging}
Our bagging is different from Breiman's original proposal in some points.
Given a sample of size $k$, Breiman's bagging makes \ul{some particular} replicates of \ul{size $k$} by random sampling \ul{with} replacement.
Our bagging aims to implicitly achieve similar effect by changing the KCS every time, which corresponds to taking \ul{all possible} replicates of \ul{size $m$} by random sampling  \ul{without} replacement.

Despite those differences, Fig.\ \ref{fig:param pc} shows that to get better results, users should set $k$ much greater than $m$.
If it is the case, many possibilities of KCS choice arise and the averaging effect of our bagging would be enhanced.
Therefore, we can gain much variance reduction that will also decrease the Hellinger distance.

\subsubsection{Computational Cost} \label{sec:complexity}
\begin{table*}[t]
\begin{center}
\caption{CPU time in seconds (average $\pm$ standard deviation over $100$-fold ICV). Bold fonts mean as in Table \ref{tbl:hellinger}.}
\label{tbl:time}
\begin{tabular}{c|cccc} \hline
Dataset							& $k$-NN				& KM					& BMP					& Fixed					\\
($d$ dimensions, $n$ training points)	& REX Kernel				& REX Kernel				& Gaussian Kernel			& Gaussian Kernel			\\
\hline
PUMS Person in WA				& 1.67e+0 $\pm$ 2.99e-2	& 6.67e+0 $\pm$ 2.13e-1	& 1.56e+0 $\pm$ 3.68e-2	& {\bf 5.19e-1} $\pm$ 3.29e-3	\\
$(d=2, n=693)$					& $(k=66, m=3)$			& $(L=35, m=4)$			& $(k=2, h=0.99)$			& $(h=0.011)$				\\
\hline
Swiss Roll						& 3.04e-1 $\pm$ 4.74e-3	& 1.26e+0 $\pm$ 1.01e-1	& 2.87e-1 $\pm$ 4.58e-3	& {\bf 1.56e-1} $\pm$ 2.76e-3	\\
$(d=3, n=100)$					& $(k=12, m=4)$			& $(L=66, m=6)$			& $(k=22, h=0.12)$			& $(h=0.056)$				\\
\hline
$3$D Road Network				& 2.01e+1 $\pm$ 3.07e-1	& 7.41e+1 $\pm$ 2.47e+0	& 1.96e+1 $\pm$ 3.11e-1	& {\bf 3.16e+0} $\pm$ 3.04e-2	\\
$(d=4, n=4348)$					& $(k=30, m=2)$			& $(L=96, m=5)$			& $(k=3, h=0.18)$			& $(h=0.009)$				\\
\hline
Letter Recognition					& 2.27e+0 $\pm$ 3.61e-3	& 8.39e+0 $\pm$ 3.13e-1	& 2.26e+0 $\pm$ 3.99e-2	& {\bf 2.06e-1} $\pm$ 3.34e-3	\\
$(d=17, n=200)$					& $(k=36, m=4)$			& $(L=68, m=20)$			& $(k=2, h=0.12)$			& $(h=0.059)$				\\
\hline
\end{tabular}
\end{center}
\end{table*}
To analyze the time complexity of the $k$-NN REX kernel, we examine \alref{alg:XKDE}.
Lines $1$--$3$ compute the $k$-NN of $n$ points of dimension $d$.
We implemented this by a naive $O(dn^2)$ time algorithm that calculates the Euclidean distances among points.
Lines $5$--$10$ compute $l$ points, each of which is generated by \eqref{eqn:REX} in $O(dm)$ time.
Therefore, the overall complexity is $O(dn^2 + dlm)$.
In a similar way, we also get the complexity of the BMP estimator as $O(dn^2 + dl)$.
The difference between them is $m$ in the second term, which is related to resampling.
Usually $m$ is small as it should be the intrinsic dimensionality plus one.
Thus, the additional computation cost for our extensions will not be so significant in most cases.

The actual computation time is shown in Table \ref{tbl:time}.
This was measured by running a single threaded C++ program on a server machine: Intel Xeon $3.07$ GHz, $64$-bit, $30$ GB RAM.
For each method, the time includes reading the dataset from a file, computing the density, synthesizing the population and writing it to a file under one particular parameter setting; excludes selecting parameters and calculating the Hellinger distance.
The table shows that for every dataset, the time of the $k$-NN REX kernel is comparable to the BMP Gaussian kernel.
This result supports the theoretical comparison of their time complexities.
Our method is approximately four times faster than the KM REX kernel in all cases, which implies that in KCS choice, our $k$-NN restriction has a certain advantage over KM's log-likelihood optimization in terms of speed.
The table also shows that the fixed Gaussian kernel is much faster than our method.
However, this is the result excluding the time of bandwidth selection.
In practice, our method has a parameter selection rule presented in Section \ref{sec:parameter sensitivity}, which enable us to find a good parameter setting within several parameter examinations.
The other methods need a kind of grid search, typically surveying tens or hundreds of parameter settings.
For fairer comparison, if the time of parameter selection is taken into consideration, our method would be the fastest.
The bottleneck of our method, and the BMP estimator, is the computation of $k$-NN as their complexities imply.
In fact, for $3$D Road Network and Letter Recognition, it consumes more than $90\%$ of overall time.
If one would like to accelerate it, a variety of nearest neighbor search techniques are available.

\section{Application to Urban Micro-Simulation} \label{sec:application}
In the last section, we have seen that the kernel density estimation generally enjoys significant improvements over sample-copying or bootstrapping approaches, given an unbiased sample, i.e., Case I.
Now we proceed to Case II, a biased sample with marginal frequencies.
Our interest in this section is whether that is yet beneficial for multi-agent simulations.

\subsection{UrbanSim} \label{sec:UrbanSim}
UrbanSim \cite{Waddell02} is one of the most advanced urban micro-simulators.
It simulates location choice of business units and households, development and pricing of land and real estate, and govermental policies involved in the study area, and can investigate their change over years in parcel level.
There are several applications in U.S. cities: Honolulu, Hawaii; Eugene-Springfield, Oregon; Detroit, Michigan; Salt Lake City, Utah; and Seattle, Washington; and in Europe: Paris, France; Brussels, Belgium; and Zurich, Switzerland.
UrbanSim requires data that describe detailed attributes of every household in the study area as a part of the initial condition for simulation.
In the United States, the U.S. Census Bureau publishes annual survey results for a $1\%$ or $5\%$ sample of households living in each area as Public Use Microdata Sample (PUMS).
The marginal distribution of each attribute of households is published by Census Summary Files (SF).
Combining them, one can generate households to run UrbanSim.

\subsection{PopGen} \label{sec:IPU}
To supply households of UrbanSim, PopGen \cite{Konduri10} is available.
This software implements Iterative Proportional Updating (IPU) \cite{Ye09} that weights each sample point to synthesize the population where the weights are determined so that the population's marginal distributions match to the given SF marginals.

\subsection{$k$-NN REX Kernel with Bias Correction} \label{sec:kNN kernel statistics}
To apply the $k$-NN REX kernel to this task, we need a little modification so that its outcome matches to SF marginals.
\alref{alg:XKDE for popsynth} is the version to do so.
In \alref{alg:XKDE for popsynth}, $F_b$ denotes the frequency in the bin $b$ and $Y_b$ denotes the subset of $Y$ whose elements are in the bin $b$.
Since UrbanSim requires integer attributes, we round the attributes of real numbers generated by this algorithm.
\begin{algorithm}[t]
\footnotesize
\caption{$k$-NN REX Kernel with Bias Correction}
\label{alg:XKDE for popsynth}
\begin{algorithmic}[1]
\Require
\Statex $X = \set{\x_1, \dots, \x_n} \subset \R^d$ \Comment{PUMS household microdata}
\Statex $F = \set{F_b \in \N \mid b \in B}$ \Comment{SF household marginal frequencies}
\Statex $k \in \set{0, \dots, n}$ \Comment{Neighborhood size}
\Statex $m \in \set{1, \dots, k+1}$ \Comment{KCS size}
\Statex $l \in \N$ \Comment{Number of households}
\Ensure
\Statex $Y = \set{\y_1, \dots, \y_l} \subset \R^d$ \Comment{Synthetic households}
\ForAll{$\x \in X$}
\State calculate the $k$-NN of $\x$, say $X(\x, k)$
\EndFor
\State $Y \gets \emptyset$
\Repeat
\State find the most vacant bin $b^* = \argmax_b F_b - \card{Y_b}$
\If{$X_{b^*} \ne \emptyset$}
\State draw $\x_1$ at random from $X_{b^*}$
\Else
\State generate $\x_1$ by sampling from the uniform distribution on $b^*$
\State calculate the $k$-NN of $\x_1$
\EndIf
\State draw $\x_2,\dots,\x_m$ $(\x_i \ne \x_j)$ from $X(\x_1, k)$ at random
\State generate $\y$ by \eqref{eqn:REX}
\State $Y \gets Y \cup \set{\y}$
\If{$\exists b \in B: \card{Y_b} > F_b$}
\State remove a point from $Y_b$ at random
\EndIf
\Until{$\card{Y} = l$}
\State \Return $Y$
\end{algorithmic}
\end{algorithm}

\subsection{Experimental Settings} \label{sec:application settings}
We used the PUMS $1\%$ sample of households in Seattle in $2000$ and extracted six attributes from them: age of head, annual income, number of people, number of kids, race, number of workers.
PopGen's IPU and our method generated $258{,}469$ households (the number of households in Seattle in $2000$).
With the households generated by each method, we ran UrbanSim from $2000$ to $2012$.
We evaluated the Hellinger distance between the simulated households and the actual ones of PUMS $5\%$ sample\footnote{In evaluation, the sampling bias on the PUMS was corrected by using PUMS weights.}.
PopGen $1.1$ was used as a baseline, but note that our settings do not involve any person attribute.
We used UrbanSim $4.4.0$ and kept all settings, other than initial households, their default values in the demo project file ``seattle\_parcel\_default.xml''.
We used $k=50$ and $m=7$ for our method, according to our parameter selection rule in Section \ref{sec:parameter sensitivity}.

\subsection{Results and Discussions} \label{sec:results and discussion}
\begin{table}[t]
\begin{center}
\caption{Hellinger distance between synthesized households and PUMS (average $\pm$ standard deviation over $10$ trials). Bold fonts mean as in Table \ref{tbl:hellinger}.}
\label{tbl:hellinger3}
%\small
\begin{tabular}{ccc} \hline
Year						& $k$-NN REX Kernel		& PopGen's IPU			\\
\hline
2000 (before UrbanSim)	& {\bf 3.94e-1} $\pm$ 1.99e-2	& 4.66e-1 $\pm$ 2.13e-2	\\
2012 (after UrbanSim)		& {\bf 4.04e-1} $\pm$ 2.21e-2	& 4.79e-1 $\pm$ 2.52e-2	\\
\hline
\end{tabular}
\end{center}
\end{table}
The result is shown in Table \ref{tbl:hellinger3}.
Before simulation, the $k$-NN REX kernel's households are closer to the PUMS $5\%$ sample than IPU's.
This tendency remains through the simulation, resulting in a better prediction.
We guess this is because the diversity of household attributes is better reproduced by our method.
IPU simply copies the given sample to synthesize the population, which never includes households that exist in a real-world city but are not present in the PUMS $5\%$ sample.
In contrast, the $k$-NN REX kernel estimates the density of the underlying population and resamples in a probabilistic manner, which can generate households that are not present in the given sample.

\section{Conclusions} \label{sec:conclusions}
In this paper, we proposed the $k$-nearest neighbor crossover kernel for population synthesis.
We showed that our method outperforms, in accuracy and speed, the conventional fixed kernel, the Breiman-Meisel-Purcell variable kernel and the Kimura-Matsumura crossover kernel.
We gave a parameter selection rule and a rationale for our method.
We demonstrated the usefulness of our method for a household synthesis task in urban simulation.

For future work, we plan to apply the proposed method to other real-world tasks, which will reveal the generality of the method.
Especially, oversampling for imbalanced data would be a fruitful application.
Another direction is to further interchange knowledge between kernel density estimation and genetic algorithm.
Theories developed in the former community may help to understand why genetic algorithms work.
Algorithms developed in the latter community can provide how to calculate large-scale estimators that are theoretically too complex to derive.

\appendices
\section{Derivation of Equation \eqref{eqn:k-NN MLE}} \label{sec:derivation}
Loftsgaarden \& Quesenberry \cite{Loftsgaarden65} showed that in the consistency analysis of their $k$-NN density estimator, if $k$ is chosen as a non-decreasing sequence of positive integers such that $k \to \infty$ and $k/n \to 0$ as $n \to \infty$, then the $k$-NN ball tends to contain arbitrarily many points while its volume converges to zero in probability as $n$ increases.
Our analysis is motivated by their result.
However, to simplify the situation, we will evaluate the analytic covariance matrix (instead of the sample one) of a random variate on a ball whose radius deterministically goes to zero.
Thus, this is not a rigorous proof but still gives an insight into the asymptotic behavior of the sample covariance of $k$-NN points in our REX kernel as $n \to \infty$.
Assume the density $f$ is continuously differentiable and has a non-zero value at the point of consideration.

We consider the first-order Taylor series approximation to the density around any point $\x \in \R^d$ to which the $k$-NN converges.
Since the covariance matrix is translation invariant, we can fix $\x = \bs 0$ without loss of generality:
$$
f(\y) = f(\bs 0) +f'(\bs 0) \y + O(\y^2).
$$
where $f' = (\nabla f)^\T = \paren{\frac{\partial f}{\partial y_1}, \dots, \frac{\partial f}{\partial y_d}}$.

Integration over a $d$-ball $B$ with radius $\delta$ centered at the origin is written by
\begin{multline*}
\int_{B} \phi(\y) \diff \y = \int_0^\delta \int_0^{2 \pi} \int_0^\pi \dots \int_0^\pi\\
\phi(\Phi(\theta_1, \dots, \theta_{d-1}, r)) J(\theta_1, \dots, \theta_{d-1}, r)\\
\diff \theta_1 \dots \diff \theta_{d-1} \diff r
\end{multline*}
where
\begin{multline*}
\Phi(\theta_1, \dots, \theta_{d-1}, r) =\\
(r \cos \theta_1, r \sin \theta_1 \cos \theta_2, \dots, r (\prod_{i=1}^{j-1} \sin \theta_i) \cos \theta_j,\\
 \dots, r (\prod_{i=1}^{d-1} \sin \theta_i) \cos \theta_{d-1}, r \prod_{i=1}^{d-1} \sin \theta_i)
\end{multline*}
and
$$
J(\theta_1, \dots, \theta_{d-1}, r) = r^{d-1} \prod_{i=1}^{d-2} (\sin \theta_i)^{d-1-i}.
$$

To calculate the integral, there are useful formulas for definite integrals of trigonometric functions: for any non-negative integers $p,q$,
\begin{equation*}
\begin{split}
\int_0^{\pi} \sin^p \theta \cos^q \theta \diff \theta &=
\begin{cases}
\Beta \paren{\frac{p+1}{2}, \frac{q+1}{2}} & \mbox{($q$ even)},\\
0                                                      & \mbox{($q$ odd)},
\end{cases}\\
\int_0^{2 \pi} \sin^p \theta \cos^q \theta \diff \theta &=
\begin{cases}
2 \Beta \paren{\frac{p+1}{2}, \frac{q+1}{2}} & \mbox{($p, q$ even)},\\
0                                                         & \mbox{(otherwise)}
\end{cases}
\end{split}
\end{equation*}
where
$$
\Beta(x, y) = \frac{\Gamma(x) \Gamma(y)}{\Gamma(x+y)}
$$
is the beta function and $\Gamma$ is the gamma function which satisfies: for any non-negative integer $r$,
\begin{equation*}
\begin{split}
\Gamma(r + 1) &= r!,\\
\Gamma(r + \frac{1}{2}) &= \frac{(2r - 1)!!}{2^r} \sqrt{\pi} = \frac{[2r - 1]!}{2^{[2r-1]} [r-1]!} \sqrt{\pi}
\end{split}
\end{equation*}
with $[x] = \max(x, 0)$.
Using these formulas and the property that $\int_B \phi(\y) \diff \y = 0$ for any odd function $\phi(-\y) = -\phi(\y)$,
we get the following equations:
\begin{equation*}
\begin{split}
\int_{B}        \diff \y &= V \delta^d,\\
\int_{B} \y    \diff \y &= \bs 0,\\
\int_{B} \y^2 \diff \y &= \frac{V \delta^{d+2}}{d + 2} \bs I,\\
\int_{B} \y^3 \diff \y &= \bs 0
\end{split}
\end{equation*}
where
\begin{equation*}
\begin{split}
V     &= \frac{\pi^{d/2}}{\Gamma(\frac{d}{2} + 1)} \ \ \  \, \qquad \mbox{(volume of unit ball)},\\
\y^2 &= \y \otimes \y^\T                       \qquad \quad \ \ \ \mbox{($d \times d$ matrix)},\\
\y^3 &= \y \otimes \y^\T \otimes \y       \qquad \mbox{($d \times d \times d$ tensor)}.
\end{split}
\end{equation*}

Let us consider the density restricted to $B$:
$$
g(\y) = f(\y| \y \in B) = \frac{1}{z} f(\y)
$$
where
\begin{equation*}
\begin{split}
z &= \int_{B} f(\y) \diff \y\\
  &\approx \int_{B} f(\bs 0) + f'(\bs 0) \y \diff \y\\
  &= f(\bs 0) \int_{B} \diff \y + f'(\bs 0) \int_{B} \y \diff \y\\
  &= f(\bs 0) \cdot V \delta^d + f'(\bs 0) \cdot \bs 0\\
  &= f(\bs 0) V \delta^d.
\end{split}
\end{equation*}

Suppose $\y$ is a random variate with density $g$.
The mean of $\y$ is
\begin{equation*}
\begin{split}
\E[\y] &= \int_{B} g(\y) \y \diff \y\\
         &= \frac{1}{z} \int_{B} f(\y) \y \diff \y\\
         &\approx \frac{1}{z} \int_{B} f(\bs 0) \y + \paren{f'(\bs 0) \y} \y \diff \y\\
         &= \frac{1}{z} \paren{ f(\bs 0) \int_{B} \y \diff \y + \int_{B} \y^2 \diff \y \nabla f(\bs 0)}\\
         &= \frac{1}{z} \paren{f(\bs 0) \cdot \bs 0 + \frac{V \delta^{d+2}}{d + 2} \bs I \cdot \nabla f(\bs 0)}\\
         &= \frac{1}{f(\bs 0) V \delta^d} \cdot \frac{V \delta^{d+2}}{d + 2} \nabla f(\bs 0)\\
         &= \frac{\delta^2}{(d + 2)} \frac{\nabla f}{f}(\bs 0).
\end{split}
\end{equation*}

The mean of $\y^2 = \y \otimes \y^\T$ is
\begin{equation*}
\begin{split}
\E[\y^2] &= \int_{B} g(\y) \y^2 \diff \y\\
            &= \frac{1}{z} \int_{B} f(\y) \y^2 \diff \y\\
            &\approx \frac{1}{z} \int_{B} f(\bs 0) \y^2 + \paren{f'(\bs 0) \y} \y^2 \diff \y\\
            &= \frac{1}{z} \paren{f(\bs 0) \int_{B} \y^2 \diff \y + f'(\bs 0) \int_{B} \y^3 \diff \y}\\
            &= \frac{1}{z} \paren{f(\bs 0) \cdot \frac{V \delta^{d+2}}{d + 2} \bs I + f'(\bs 0) \cdot \bs 0}\\
            &= \frac{1}{f(\bs 0) V \delta^d} \cdot \frac{f(\bs 0) V \delta^{d+2}}{d + 2} \bs I\\
           &= \frac{\delta^2}{d + 2} \bs I.
\end{split}
\end{equation*}

Finally, the covariance of $\y$ is
\begin{equation*}
\begin{split}
\C[\y] &= \E[\y^2] - \E[\y]^2\\
        &= \frac{\delta^2}{d + 2} \bs I - \paren{\frac{\delta^2}{(d + 2)} \frac{\nabla f}{f}(\bs 0)}^2\\
        &= \frac{\delta^2}{d + 2} \bs I - \frac{\delta^4}{(d + 2)^2} \frac{\nabla f f'}{f^2}(\bs 0).
\end{split}
\end{equation*}

This result implies that our $k$-NN REX kernel, more generally, generic $k$-NN crossover kernels such that their crossovers are compliant to Kita's preservation of statistics, has the same order of bandwidth as the BMP estimator.
Especially, if we choose the KCS size of the $k$-NN REX kernel as $m = k + 1$ and the constant multiplier of the BMP estimator as $h = 1/\sqrt{d+2}$, then both asymptotically coincides.
Consequently, the $k$-NN REX kernel can be considered as a matrix bandwidth extension of the BMP estimator.

\section{Visual Comparison} \label{sec:visual comparison}
\begin{figure}[htb]
\hspace{-8mm}
\begin{tabular}{cc}
\ig[0.5]{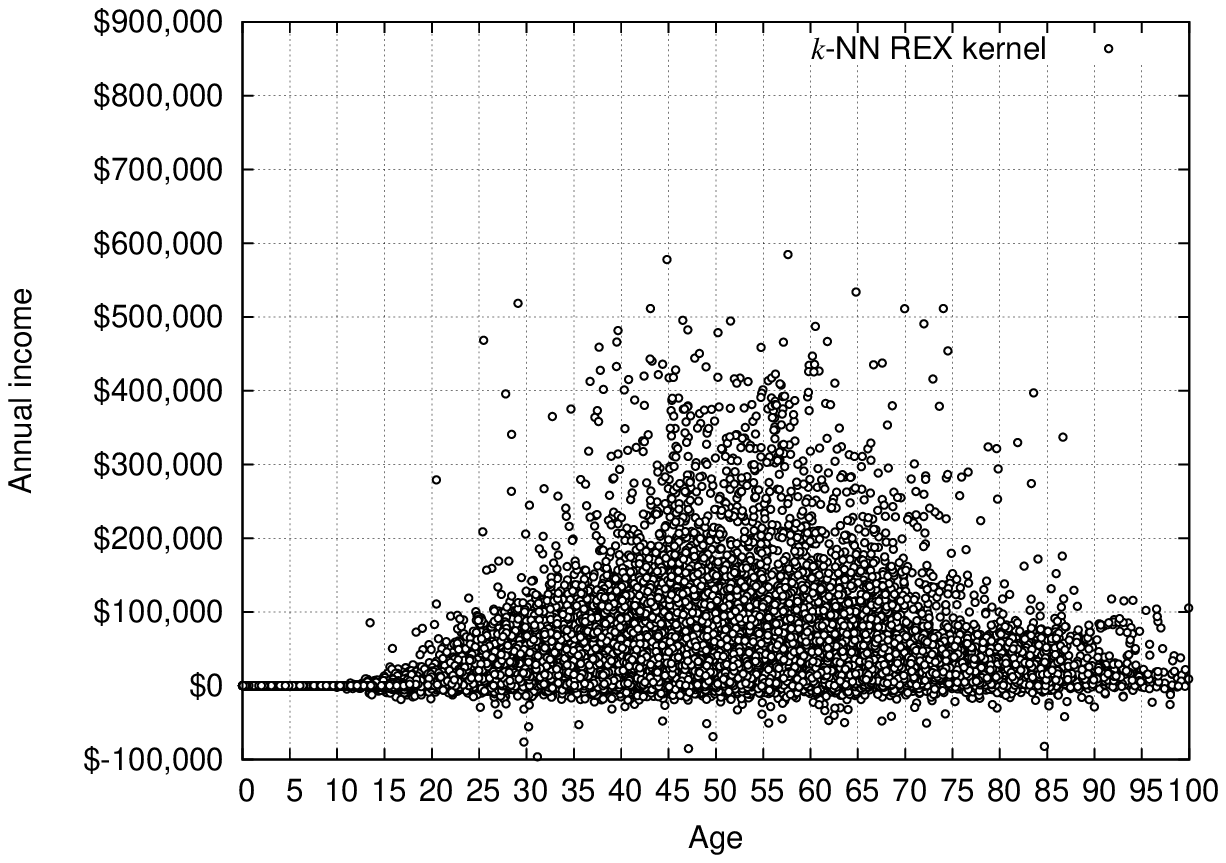}
\ig[0.5]{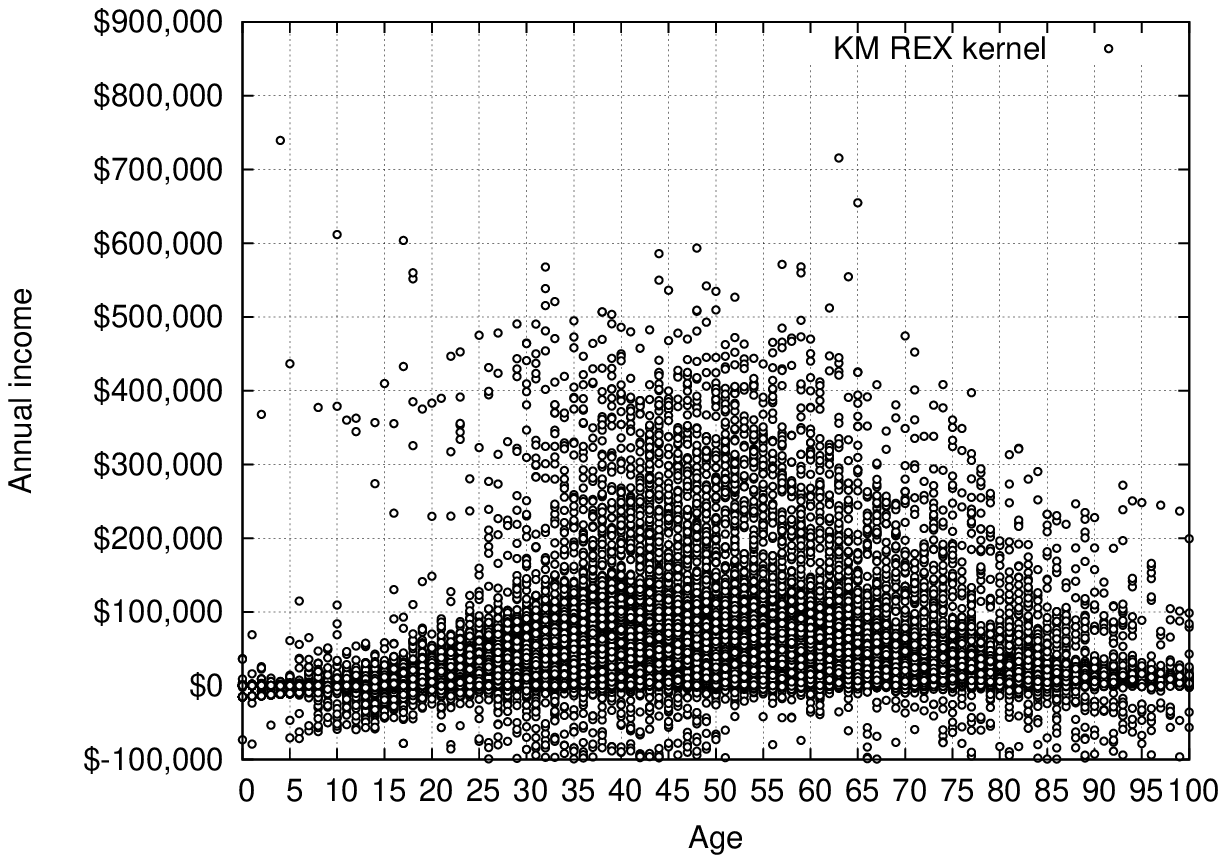}\\
\ig[0.5]{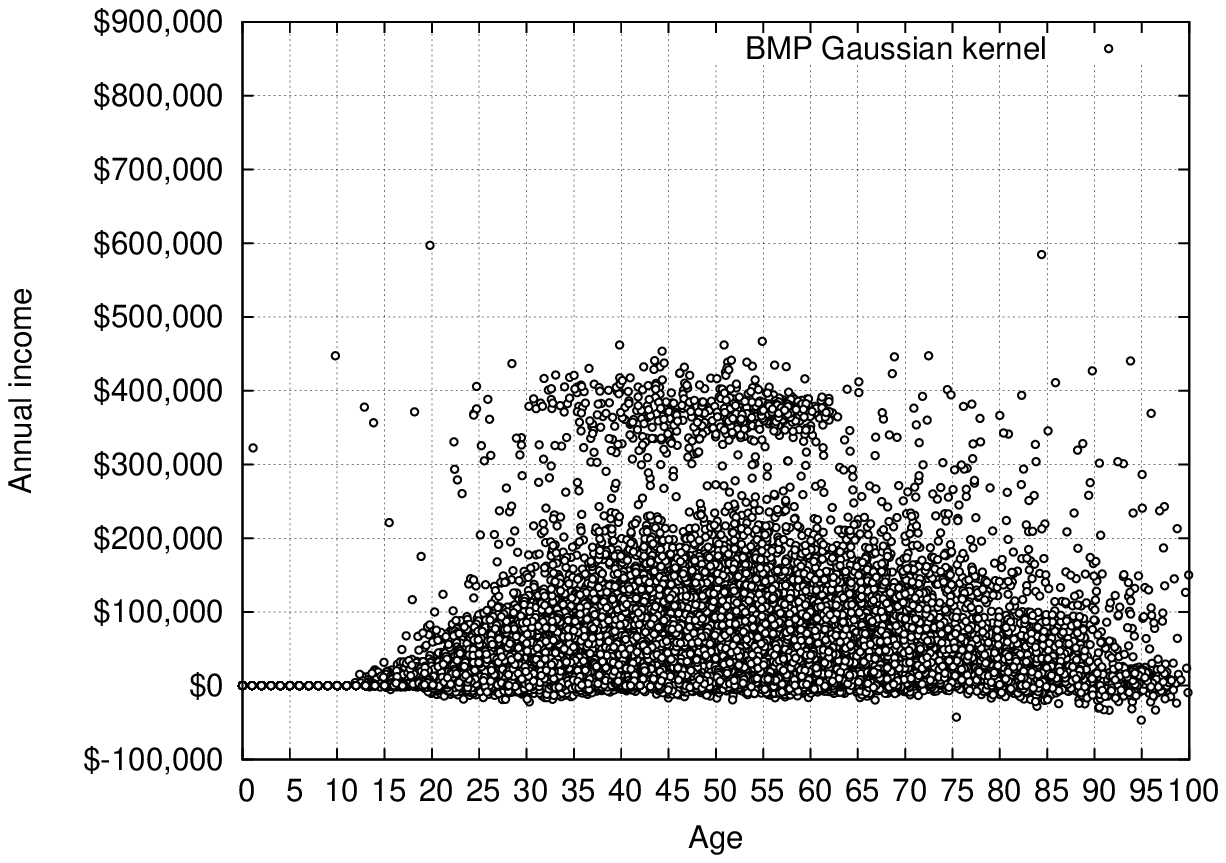}
\ig[0.5]{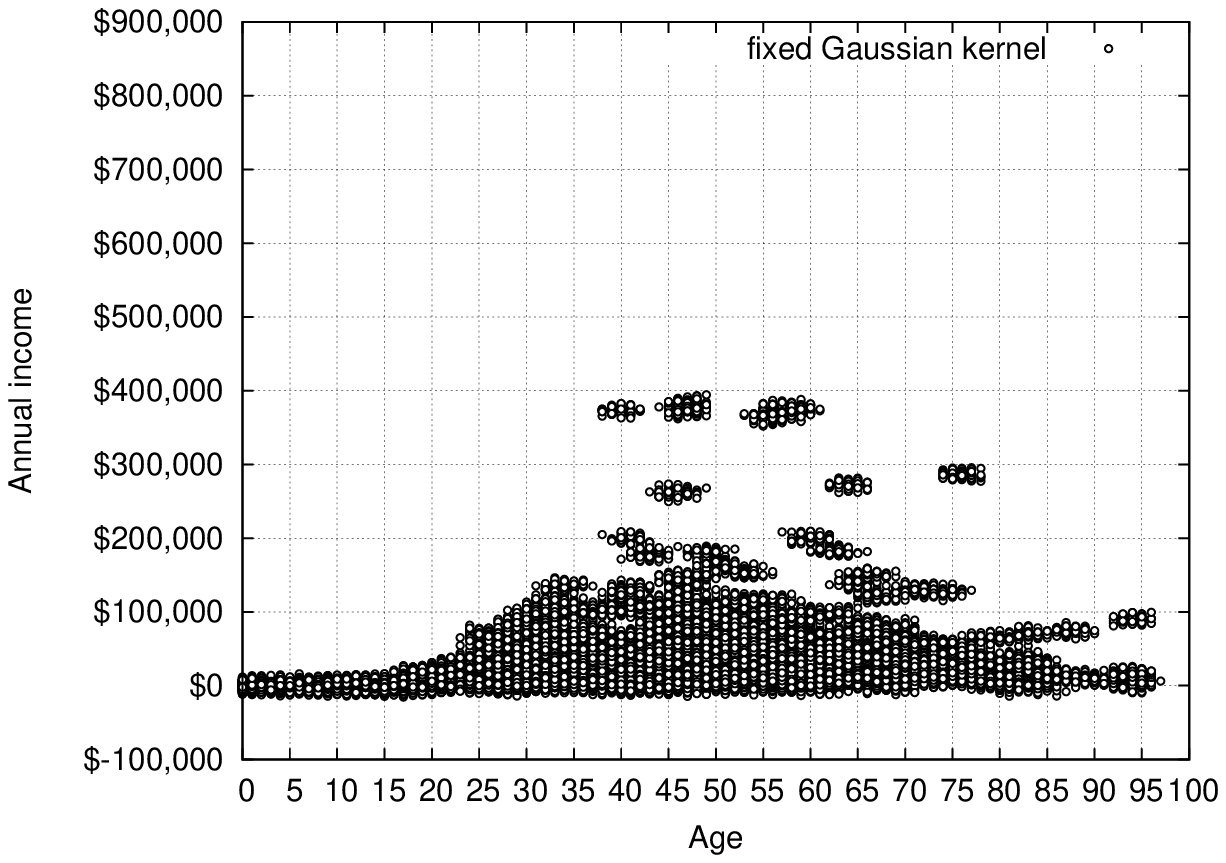}\\
\ig[0.5]{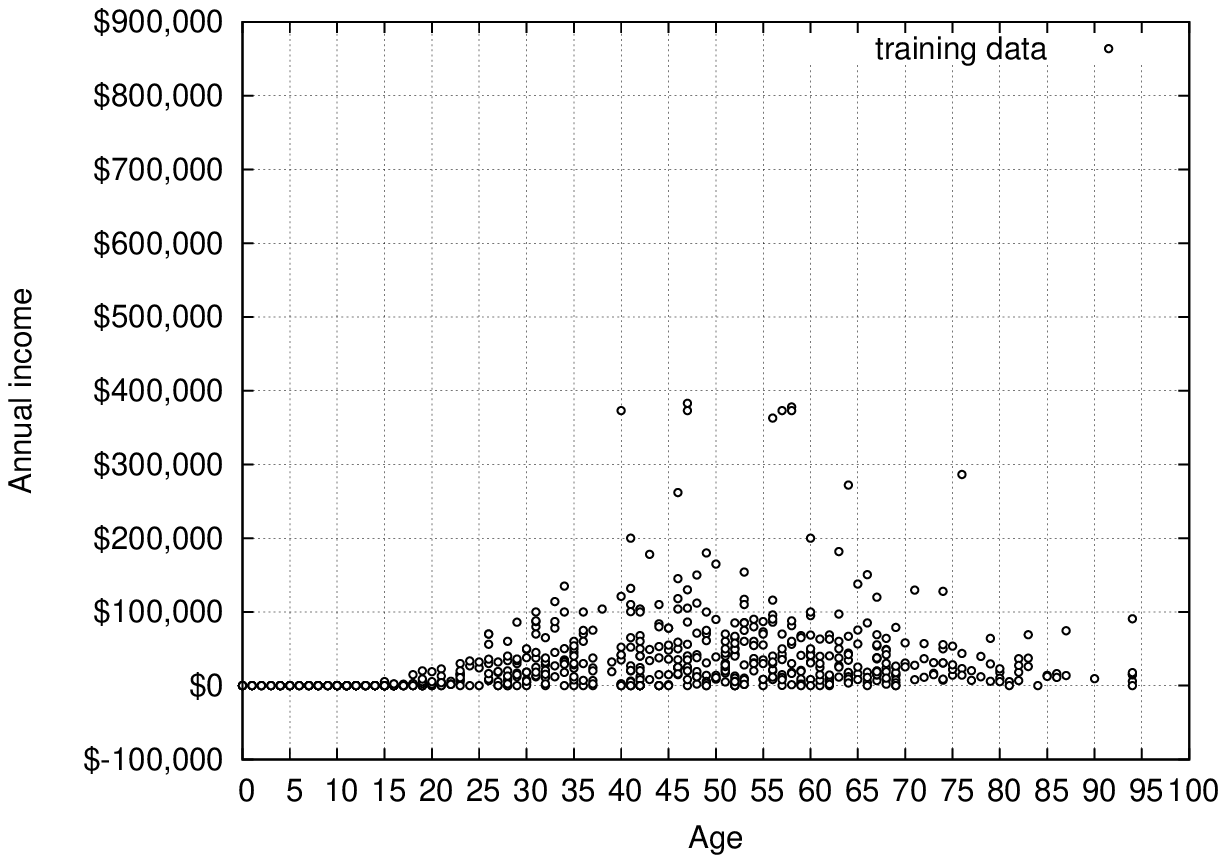}
\ig[0.5]{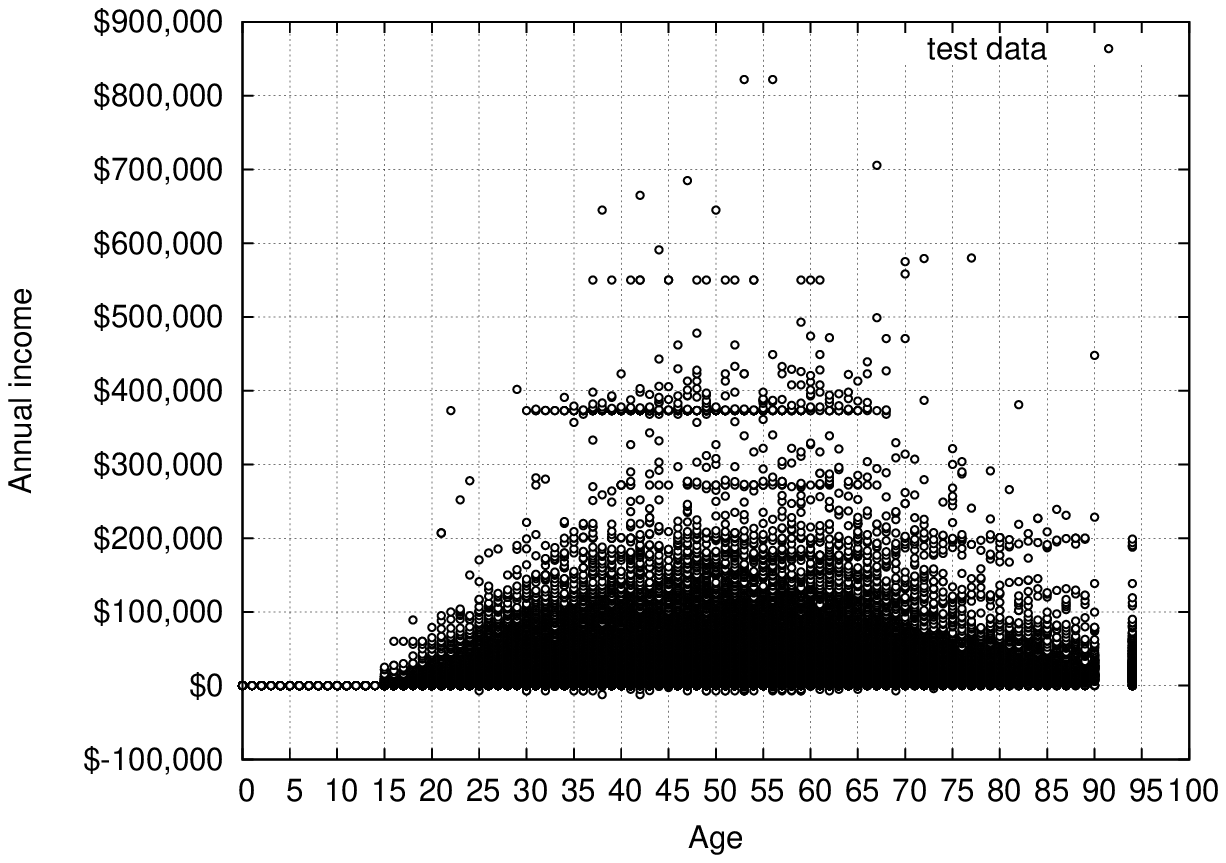}\\
\mpg{
\caption{Synthetic populations, training data and test data for PUMS Person in WA.}
\label{fig:population}
}
\end{tabular}
\end{figure}

Figure \ref{fig:population} shows the distribution of synthetic populations for PUMS Person in WA.
These plots intuitively support the correctness of our numerical evaluation presented in Table \ref{tbl:hellinger}.

%\section*{Acknowledgment}



\bibliographystyle{IEEEtran}
\bibliography{ref}
%\begin{thebibliography}{1}
%\end{thebibliography}

\end{document}